\documentclass{article} % For LaTeX2e
\usepackage{iclr2021_conference,times}

\usepackage{amsmath,amsfonts,bm}

\def\eqref#1{equation~\ref{#1}}
\def\1{\bm{1}}

\def\mI{{\bm{I}}}

\DeclareMathAlphabet{\mathsfit}{\encodingdefault}{\sfdefault}{m}{sl}
\SetMathAlphabet{\mathsfit}{bold}{\encodingdefault}{\sfdefault}{bx}{n}

\usepackage{hyperref}
\usepackage{url}
\usepackage{graphicx}
\usepackage{subcaption}
\usepackage{xcolor}
\usepackage{amsmath}
\usepackage{amsfonts}
\usepackage{wrapfig}
\usepackage{fancyhdr}

\title{LoRD: \textbf{Lo}w \textbf{R}ank \textbf{D}ecomposition of \\ monolingual code LLMs for one-shot compression}

\author{Ayush Kaushal \\
  Universit\'{e} de Montr\'{e}al, Nolano AI \\
\texttt{ayush@nolano.ai}
\And
  Tejas Vaidhya\\
  Mila, Universit\'{e} de Montr\'{e}al, Nolano AI \\
  \texttt{tejas@nolano.ai} 
\And
 Irina Rish\\
Mila, Universit\'{e} de Montr\'{e}al, Nolano AI\\
 \texttt{irina@nolano.ai}
}

\newcommand{\methodname}[1]{\textsc{Lo}w \textsc{R}ank \textsc{D}ecomposition (LoRD) {#1}}
\newcommand{\methodnameshort}[1]{LoRD{#1}}

\iclrfinalcopy % Uncomment for camera-ready version, but NOT for submission.
\begin{document}

\maketitle

\begin{abstract}

Low Rank Decomposition of matrix - splitting a large matrix into a product of two smaller matrix offers a means for compression that reduces the parameters of a model without sparsification, and hence delivering more speedup on modern hardware. Moreover, unlike quantization, the compressed linear layers remain fully differentiable and all the parameters trainable, while being able to leverage the existing highly efficient kernels over floating point matrices. We study the potential to compress Large Language Models (LLMs) for monolingual Code generation via \methodname{} and observe that ranks for the linear layers in these models can be reduced by upto 39.58\% with less than 1\% increase in perplexity. We then use \methodnameshort{} to compress StarCoder 16B to 13.2B parameter with no drop and to 12.3B with minimal drop in HumanEval Pass@1 score, in less than 10 minutes on a single A100. The compressed models speeds up inference by up to 22.35\% with just a single line of change in code over huggingface's implementation with pytorch backend. \methodnameshort{} models remain compatible with state of the art near-lossless quantization method such as SpQR, which allows leveraging further compression gains of quantization. Lastly, QLoRA over \methodnameshort{} model further reduces memory requirements by as much as 21.2\% over vanilla QLoRA while offering similar gains from parameter efficient fine tuning. Our work shows \methodname{} as a promising new paradigm for LLM compression. \footnote{We will release \methodnameshort{}Coder at https://huggingface.co/nolanoAI}

\end{abstract}

\section{Introduction}

Code LLMs have become an integral component of Copilots that boost developer productivity \citep{copilot_productivity_gains} and in LLM based agents \citep{survey_on_llm_agents}. These Code LLMs are as large as 34 Billion parameters for the publicly available models \cite{codellama_paper} and more than 175 Billion parameter for closed source ones \cite{humaneval_paper}. There is not only a pressing need for reducing model size and running models at a lower cost, but also for increasing the inference speed. The latter is especially significant for Copilot based applications.

Recently, several methods have been proposed to compress and speed up inference of LLMs. Quantization \citep{GPTQ_paper_elias_frantar_dan_alistarh, SpQR_tim_dettmers} reduces the number of bits required per weight parameter of LLM by lowering the precision, and has shown significant model compression as well as speedups in low-batch decoding phases of LLMs \cite{squeezellm_paper}. Quantization has also been shown to generalize well to quantized models \cite{PanGu-Coder2}. Pruning \citep{wanda_zico_kolter, sparsegpt_elias_frantar_dan_alistarh} has offered another means of compression by removing connections from the neural network and hence sparsifying the weight matrices of the neural networks. Distillation \cite{minillm_distillation_microsoft, gkd_distillation_deepmind, impossible_distillation_yejin_choi} method enables one to train a smaller model using a larger teacher model for supervision. While quantization and pruning methods that do not require re-training are viable means of compressing the model, distillation involves a significant amount of compute for retraining a smaller LLM, often from scratch. Here, we consider another compression paradigm of \methodname{}, that does not require expensive retraining as in the case of distillation and covers up several deficiencies of the quantization and pruning compression method.

% Motivating the benfits of LoRD.
Low Rank Decomposition factorizes a dense matrix of a neural network as a product of two smaller dense matrices. The \methodnameshort{} model can leverage the highly optimized floating-point dense matrix multiplication kernels \citep{cuda, blas_citation} that have been written over modern hardware. In contrast, quantized models require specialized kernels to be written, often different for each hardware backend in order to enable fast inference. Moreover, the neural network remaining fully-differentiable and all the parameters remaining trainable even after compression, unlike quantization. The LoRA \cite{lora_paper} layers of tuned models are also easier to merge back into floating point matrices compared to the quantized ones.

Pruned models produce sparse matrix weights in the neural network. Matrix multiplication over sparse matrices is much slower than the resulting dense matrices in \methodnameshort{} on most GPUs. Dense matrices, in addition avoid representation format overhead that sparse matrices incur from parameter reduction \footnote{This overhead in sparse matrix occurs from having to store indices/bitmasks to indicate which values are present and not. This can be very significant at low levels of sparsity. PyTorch's sparse formats (CSR, CSC, COO) all store indices at int64 format, and for moderate levels of sparsity ($<$50\%), the sparse matrix takes up more space than a dense matrix with zero-ed out values.} and often requires specialized kernels for reducing this overhead \cite{SpQR_tim_dettmers}. Dense matrix multiplication is also easier to implement than sparse matrix multiplication, especially over quantized models.

% Motivating why LoRD would be possible for LLMs. For this, cite works claiming low-dimensionality of ranks in neural networks. Then show that this works has been used to compress Smaller Language Models, but only for task specfic or via distillation.

Several previous works have attempted to apply matrix decomposition methods like SVD, Tucker or Kronecker decomposition for compression \citep{Compressing_Pre-trained_Language_Models_by_Matrix_Decomposition_yoav_goldeberg, kroneckerbert, Kronecker_Decomposition_for_GPT_Compression}. However, these have been limited to small language models like Bert \citep{bert_paper_devlin_et_al} and GPT2 \citep{gpt2_paper}, and have shown success only on narrow task-specific use cases or after retraining, often only with teacher-guided distillation supervision. These works have observed that weight matrices are not low rank and adapt methods like Singular Value Decomposition for data-aware decomposition of weights \citep{DRONE_Dataaware_Lowrank_Compression_for_Large_NLP_Models, Language_model_compression_with_weighted_lowrank_factorization, Compressing_transformers_features_are_lowrank_but_weights_are_not}.

We, adapt these approaches for Large Language Models (Billion+ Parameters) over python code, and show that these models can be low-rank decomposed to compress and speed up inference without the need for retraining with little to no performance degradation. We study low-rank decomposition across two families of code LLMs - StarCoder and CodeGen (\S\ref{sec:Code_LLMs_are_Low_Rank_Decomposable}) for varying parameter sizes and establish the potential for reducing rank of models through decomposition. We then study these trends across different kinds of linear layers in a transformer block and observe the potential for upto 39.58\% rank reduction with less than 1\% change in perplexity.

We propose various considerations for compressing the models and to achieve inference speedup on GPUs (\S\ref{sec:compression_and_speedup_by_decomposition::subsec:Achieving_compression_and_inference_speedup}). Using these, we achieve compression of the StarCoder 16B model offering 31.67 HumanEval \cite{humaneval_paper} Pass@1 score down to 13.2B parameter with similar performance of 31.57 HumanEval and down to 12.3B parameter with 29.22 HumanEval score (\S\ref{sec:compression_and_speedup_by_decomposition::subsec:Performance_of_compressed_models}). \methodnameshort{} models, offer an inference speedup of as high as 22.35\% with just one line of change in huggingface's (\S\ref{sec:compression_and_speedup_by_decomposition::subsec:Speedup_from_LoRD}).

These \methodnameshort{} models can be further compressed via the near-lossless quantization method of SpQR \cite{SpQR_tim_dettmers} to reduce it's precision to 8 and 4 bits without any further reduction in HumanEval performance (\S\ref{sec:combining_lord_with_quantization_and_lora::subsec:quantization}). Finally, these decomposed models also reduce the memory requirements of adapter finetuning by 21.2\% over QLoRA (\S\ref{sec:combining_lord_with_quantization_and_lora::subsec:Parameter_Efficient_tuning_of_lord}).

\section{Code LLMs are Low Rank Decomposable}\label{sec:Code_LLMs_are_Low_Rank_Decomposable}

\subsection{Background}

Let an linear layer $L$ of an LLM $M$ with weight $W \in \mathbb{R}^{d_1\times d_2}$ and bias $b \in \mathbb{R}^{d_1\times 1}$. Let $d_{min} = minimum(d_1, d_2)$ and $d_{max} = maximum(d_1, d_2)$

A \textbf{Low Rank Decomposition} or Low Rank Factorization of a layer $L$ would give us a new layer $\Tilde{L}$ with two weight matrices $A \in \mathbb{R}^{r \times d_2}$ and $B \in \mathbb{R}^{d_1 \times r}$, and a bias $\Tilde{b} \in \mathbb{R}^{d_1 \times 1}$, where $r<<d_{min}$ such that for a $n$ batch of input vectors $X \in \mathbb{R}^{d_2 \times n}$ the batch of output vectors $Y \in \mathbb{R}^{d_1 \times n}$ is,

\begin{equation}
\label{linear_decomposition_equation}
Y = \Tilde{L}(X) = B A X + \Tilde{b} \approx L(X) = W X + b
\end{equation}

Singular Value Decomposition (SVD) offers the best $r$-rank approximation of matrix $W\in \mathbb{R}^{d_1 \times d_2}$. First $W$ can be decomposed as $W = USV^T$, where $U \in \mathbb{R}^{d_1 \times d_2}$ and $V \in \mathbb{R}^{d_2 \times d_2}$ are orthogonal matrix and $S \in \mathbb{R}^{d_1 \times d_2}$ is a diagonal matrix with entries in decreasing order. Then, by taking top-k rank, we can decompose $W$ as a product of two low ranked matrices $W \approx BA$ as follows%, where $B=U_{:, :r}S_{:r, :r} \in \mathbb{R}^{d \times r}$ and $A=V_{:r, :}  \in \mathbb{R}^{d \times r}$

\begin{equation}
W = \underbrace{(U_{:, :r}S_{:r, :r})}_{B \in \mathbb{R}^{d_1 \times r}} \underbrace{(V_{:r, :})}_{A \in \mathbb{R}^{r \times d_2}}
\end{equation}

where ${}_{:a, :b}$ denotes a slice operation over a matrix that gives its first $a$ rows and $b$ columns.

Eigendecomposition is another decomposition method applicable to symmetric matrices. We can represent the eigendecomposition of a symmetric matrix $W \in \mathbb{R}^{d_1 \times d_1}$ as $W = Q\Lambda Q^T$. Here $Q \in \mathbb{R}^{d_1 \times d_1}$ is an orthogonal matrix whose columns are the eigenvectors of $W$, and $\Lambda \in \mathbb{R}^{d_1 \times d_1}$ is a diagonal matrix whose entries are the eigenvalues of $W$ sorted in decreasing order. Similar to SVD, we can decompose $W$ as a product of two low ranked matrices $W \approx BA$ by retaining only the largest $r$ eigenvalues (and corresponding eigenvectors) as follows:

\begin{equation}
W = \underbrace{(Q_{:, :r}\Lambda_{:r, :r})}_{B \in \mathbb{R}^{d_1 \times r}} \underbrace{(Q_{:r, :}^T)}_{A \in \mathbb{R}^{r \times d_1}}
\end{equation}

Since $Q$ is orthonormal and the eigenvalues $\Lambda$ is sorted in descending order, $Q_{:, :r} Q_{:, :r}^T \approx \mI$ where $\mI$ is identity matrix of dimension $d_1$.

While SVD gives the optimal low-rank decomposition of matrix, in terms of Frobenius norm, but does not take input and output data distribution into account. Approaches like weighted SVD \citep{Language_model_compression_with_weighted_lowrank_factorization} and SVD over both weight and data \citep{DRONE_Dataaware_Lowrank_Compression_for_Large_NLP_Models} have been proposed but are prohibitively expensive to scale to larger models for their requirement of backpropagation over calibration dataset. SVD over very large weight matrices is also very computationally expensive. So, we instead leverage the observation that activations in transformers are low-ranked \citep{rank_diminishing_in_deep_neural_networks_michael_jordan} and adapt the more heuristically driven approach of Atomic Feature Mimicking (AFM) \citep{Compressing_transformers_features_are_lowrank_but_weights_are_not} that creates low rank matrices conditioned on a small amount of calibration data. Specifically, consider the eigen-decomposition of Covariance over $Y$ as

\begin{equation}
\mathbb{E}[yy^T] - \mathbb{E}[y]\mathbb{E}[y]^T = \Hat{Q}\Hat{\Lambda}\Hat{Q}^T
\end{equation}

Here $\Hat{Q}$ is a matrix of its eigenvectors, hence $\Hat{Q}_{:, :r} \Hat{Q}_{:, :r}^T \approx \mI$. Using this, we can write the output vector $Y$ as $Y \approx \Hat{Q}_{:, :r} \Hat{Q}_{:, :r}^T Y$. By writing $Y$ in terms of $W$, $X$ and $b$ from Equation \ref{linear_decomposition_equation}, we have:

\begin{equation}
    Y \approx \Hat{Q}_{:, :r} \Hat{Q}_{:, :r}^T W X + \Hat{Q}_{:, :r} \Hat{Q}_{:, :r}^T b 
\end{equation}

Comparing to Equation \ref{linear_decomposition_equation}, this gives us $B = \Hat{Q}_{:, :r} \in \mathbb{R}^{d_1 \times r}$, $A = \Hat{Q}_{:, :r}^T W \in \mathbb{R}^{r\times d_2}$  and $\tilde{b} = \Hat{Q}_{:, :r} \Hat{Q}_{:, :r}^T b \approx b$.
This approach is also straightforward to adapt for LLMs like LLaMa \citep{llama_paper}, Falcon \citep{falcon_paper}, CodeLLaMa \citep{codellama_paper} which do not have a bias term in the linear layer by setting $\Tilde{b}$ to zero vector.

\subsection{Experimental Settings}

We take our python calibration dataset from the stack \citep{the_stack_paper} and consider the corresponding subset of the stack smol \citep{the_stack_smol_huggignface} as validation data. We filter out those sequences which are less than 1024 tokens or 10240 characters in length. We consider CodeGen and StarCoder model family of models. CodeGen mono models are present across 350M, 2B, 6B and 16B parameters and are CodeGen models that were further trained on only python code. StarCoder 16B is the StarCoderBase 16B model further trained on only python code from the stack dataset's train split. We also consider StarCoderBase at 3B and 7B parameter sizes in StarCoder family due to the lack of their monolingual counterparts. All our experiments were performed on a single A100 GPU in under an hour for each run.

For studying the trends of increase in perplexity for a reduction in rank across difference model sizes, we set a fixed low-rank $r$ for all the layers. Later we discuss how to achieve compression and inference speedup via low-rank decomposition in \S\ref{sec:compression_and_speedup_by_decomposition}

\subsection{Change in Perplexity across Reduction in Rank}\label{sec:Code_LLMs_are_Low_Rank_Decomposable::subsec:Change_in_Perplexity_across_Reduction_in_Rank}

\begin{figure}[htbp]
  \centering
  \begin{subfigure}[b]{0.48\linewidth}
    \includegraphics[width=\linewidth]{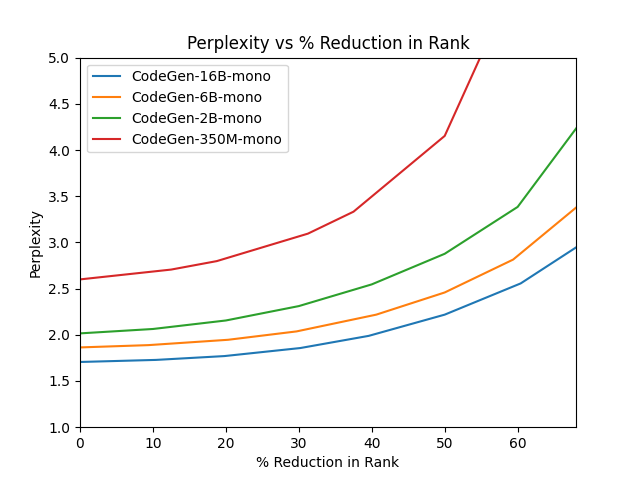}
    \caption{Perplexity vs \% Rank Reduction for CodeGen Models.}
    \label{fig:perplexity_vs_rank_reduction_codegen}
  \end{subfigure}
  \hfill
  \begin{subfigure}[b]{0.48\linewidth}
    \includegraphics[width=\linewidth]{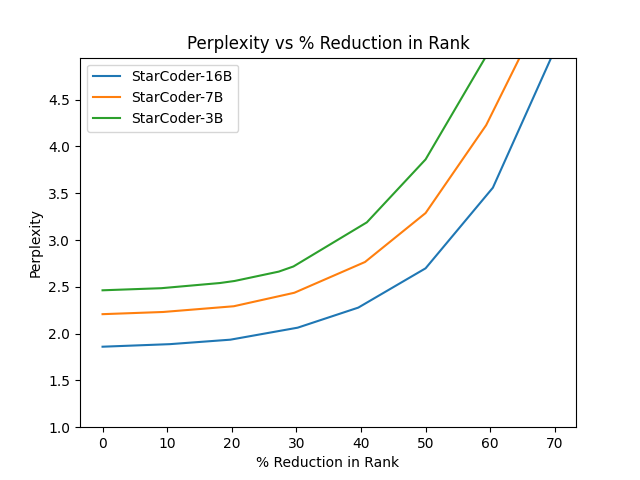}
    \caption{Perplexity vs \% Rank Reduction for StarCoder Models.}
    \label{fig:perplexity_vs_rank_reduction_starcoder}
  \end{subfigure}
  \caption{Perplexity vs \%Reduction in Rank for Different Models.}
  \label{fig:perplexity_vs_rank_reduction}
\end{figure}

Figure \ref{fig:perplexity_vs_rank_reduction_codegen} and \ref{fig:perplexity_vs_rank_reduction_starcoder} show the trends of increase in perplexity across reduction in rank of the weight matrix of CodeGen and StarCoder models. For the largest models in both families, we observe only about a 1\% increase in perplexity for 10\% reduction in rank, and upto 35\% reduction in rank for less than 10\% increase in perplexity. The smallest model, CodeGen Mono 350M, however, can only be decomposed to 35\% rank reduction for a similar drop in perplexity. We observe that the perplexity changes much slower for larger models as the \% rank reduces, and hence can be compressed mode, similar to observations in quantization and pruning \citep{train_large_then_compress}. It should be noted that for most models, more than 50\% leads to significant output quality degradation.

% \subsection{Scaling Laws}

\section{Compression and speedup through Decomposition}\label{sec:compression_and_speedup_by_decomposition}

In this section, we discuss how we adapt the \methodname{} for reducing the size of model and achieving inference speedup without a significant reduction in the output quality of the model. Following \citep{squeezellm_paper}, we assume memory bandwidth is the bottleneck for inference, and thus speedups for decoding are directly proportional to the size of the transformer model.

\subsection{Achieving compression and inference speedup}\label{sec:compression_and_speedup_by_decomposition::subsec:Achieving_compression_and_inference_speedup}

\textbf{Threshold for size reduction across rank reduction:} Consider a weight matrix $W \in \mathbb{R}^{d_1 \times d_2}$ of a transformer layer with low rank decomposed $A \in \mathbb{R}^{r \times d_2}$ and $B \in \mathbb{R}^{d_1 \times r}$. The number of parameters before and after decomposition respectively are $d_1 d_2$ and $r(d_1+d_2)$. Therefore, %the change in parameter is $r(d_1 + d_2) - (d_1 d_2)$.
if $r > \frac{d_1 d_2}{(d_1 + d_2)}$, (i.e a decomposition with small rank reduction), then the size of the model after decomposition can even be higher than the original models. Ideally, we would want the rank $r << \frac{d_1d_2}{(d_1+d_2)}$ or $r << d_{min}$.

% \redtext{AYUSH: I am introducing new notation here as well as in the previous section, and will be having more in next one. Different sections are self-contained but perhaps introducing new notation everywhere can be avoided. (Maybe not big issue for preprints)?}

\textbf{Matrix Aspect Ratio and Compression:} Let the ratio of the smaller dimension to the larger dimension of the matrix (i.e. the aspect ratio) be $\alpha=\frac{d_{min}}{d_{max}}$. For square matrix, $\alpha = 1$ and for tall or fat matrices $\alpha << 1$.
We can rewrite, the percentage change in parameters from decomposition, in terms of percent change in rank $\%\Delta r = 100*\frac{d_{min} - r}{d_{min}}\%$ and aspect ratio as:

\begin{equation}\label{equation:percent_param_reduction_in_terms_of_aspect_ratio_and_percent_rank_change}
 100*\frac{r(d_{max} + d_{min}) - d_{max} d_{min}}{d_{max} d_{min}} = 100\alpha - (1+\alpha) \%\Delta r   
\end{equation}

It should be noted that change in parameters from decomposition can either be positive (the number of parameters increased after decomposition), or negative (the number of parameters decreased after decomposition). In order to achieve model compression and consequently inference speedups, one would want a very high negative percentage change in parameters. % Based on the function, one can observe, that   order to start achieving any compression and for less reduction than that, one will observe an increase in model size, instead of reduction.

\begin{wrapfigure}{r}{0.44\textwidth}
  \centering
  \includegraphics[width=0.45\textwidth]{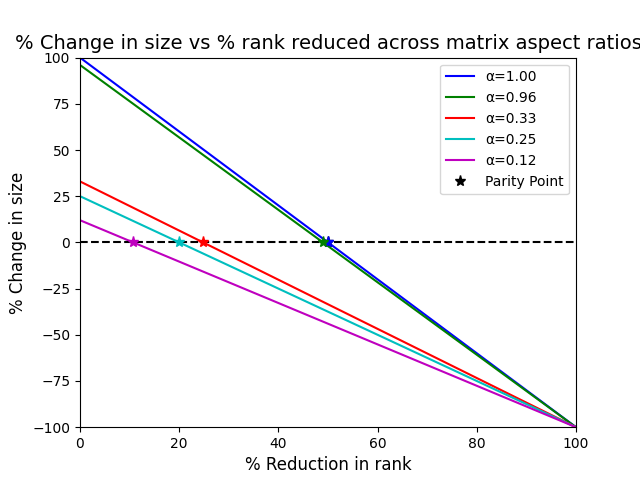}
  \caption{\label{figure:parity_point_across_aspect_ratios}Parity Point across various  aspect ratios ($\alpha$) of the different linear layers in transformers.}
\end{wrapfigure}

\textbf{Parity Point for Compression across Rank Reduction:} Using Eq. \ref{equation:percent_param_reduction_in_terms_of_aspect_ratio_and_percent_rank_change}, one can observe that little reduction in rank, may lead to increase in model parameters instead of decreasing. For instance, square matrices ($\alpha = 1$) will have 100\% increase (i.e doubling in size), then $\%\Delta r \rightarrow 0_+$ and only after the rank is reduced by more than 50\%, will the \textbf{Parity Point} of the rank reduction be reached, that offers same or lesser number of a parameter in the decomposed layer as the original matrix. This parity point for tall or fat matrices ($\alpha \rightarrow 0_+$), can be achieved with a very small percent reduction in rank and can start giving a reduction in model size. For compression to be achieved, we would want to reduce the rank by an amount to cross this parity point threshold. However, reducing the rank by a lot can degrade performance significantly. So we must take the aspect ratio into account, in order to achieve compression without much reduction in rank (and hence no significant degradation in output quality)

A transformer model had different aspect ratios across its various linear layers, $\alpha=1.00$ for output projection after attention, $\alpha=0.96$ for Multi-query attention \citep{multi_query_attention_paper} projections, $\alpha=0.25$ for typical MLP projections with intermediate expansion factor of 4 as in the original transformer and as low as $\alpha=0.12$ for the embedding and language model head projection of CodeGen 16B with 51200 vocab size. Figure \ref{figure:parity_point_across_aspect_ratios} plots the \% change in the size of the model across \% reduction in rank for matrices with different aspect ratios. For square matrices and near square matrices, a small rank reduction doubles the size of the linear layer after decomposition, and only after its parity point of 50\% reduction is the size after decomposition, the same as original matrix. By this extent of rank decomposition, the performance starts to significantly degrade, as seen in \S\ref{sec:Code_LLMs_are_Low_Rank_Decomposable::subsec:Change_in_Perplexity_across_Reduction_in_Rank}. All the previous works on smaller models, address this by retraining the model \citep{Compressing_transformers_features_are_lowrank_but_weights_are_not, DRONE_Dataaware_Lowrank_Compression_for_Large_NLP_Models, Language_model_compression_with_weighted_lowrank_factorization, Compressing_Pre-trained_Language_Models_by_Matrix_Decomposition_yoav_goldeberg}, often via knowledge distillation supervision \citep{hinton_knowledge_distillation, distilbert_paper} on specific narrow tasks. However, retraining is infeasible for larger models. Thus, we skip matrices with very high aspect ratios such as output projection or multi-query attention for decomposition. In contrast, the weights in MLP achieve parity at only 20\% rank reduction. While embeddings and LM Head can be compressed through decomposition, as they have been for smaller transformer models \citep{embedding_matrix_factorization_Adaptive_Input_Representations_for_Neural_Language_Modeling, embedding_matrix_factorization_ALBERT:_A_Lite_BERT_for_Self-supervised_Learning_of_Language_Representations}, they contribute only a very small portion of the weight of the model. So, we do not consider decomposing these matrices. In order to reduce the aspect ratio of matrices, we \textbf{group layers} with the same input vector to have the same bottleneck matrix after decomposition. Doing so, enables re-use of computation, and sharing of weights, as well as bringing the aspect ratio down to achieve compression as lower rank reduction. Candidate layers for grouping include the query, key and value projection matrices in multi-headed attention with aspect ratio reduced to $\alpha=0.33$ and the gating layer in SwiGLU \citep{swiglu_paper} with first linear linear of MLP in models like LLaMa \citep{llama_paper} with $\alpha=0.1875$.

\begin{figure}[htbp]
  \centering
  \begin{subfigure}[b]{0.48\linewidth}
    \includegraphics[width=\linewidth]{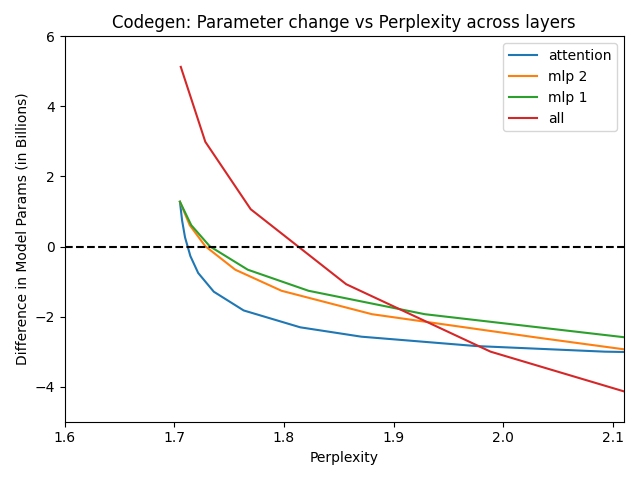}
    \caption{\label{figure:codegen_layerwise_param_reduced_vs_perplexity} CodeGen 16B.}
    
  \end{subfigure}
  \hfill
  \begin{subfigure}[b]{0.48\linewidth}
    \includegraphics[width=\linewidth]{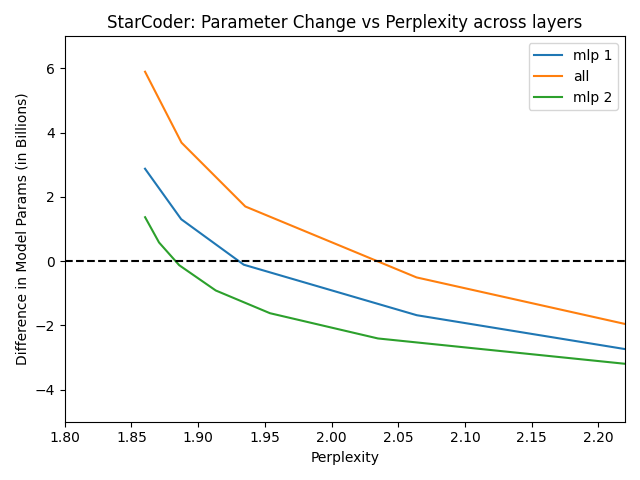}
    \caption{\label{figure:starcoder_layerwise_param_reduced_vs_perplexity} StarCoder 16B.}
  \end{subfigure}
  \caption{\label{figure:layerwise_param_reduced_vs_perplexity} Parameter Reduction vs perplexity for decomposition across various layers.}
  
\end{figure}

\paragraph{Trends across different layers in a transformer block:} In addition to considering the parity point into account for deciding which layers to decompose, we also additionally study the sensitivity of each of these layers to low rank decomposition across the large model in the two model families. Figure \ref{figure:layerwise_param_reduced_vs_perplexity} shows the increase in perplexity vs reduction in model parameters for the two models. For both models, decomposing all the linear layers achieves the parity point much later than any one of these linear layers with low aspect ratio. For CodeGen, the attention weight matrix (query, key and values projection) offers least increase in perplexity for the biggest drop in parameter count, make this layer the most suitable candidate to be decomposed. It shows less than 1\% increase in perplexity even after 39.58\% rank reduction. We observe the mlp 2 (downscaling mlp) to be a better candidate for decomposition than mlp 1 (upscaling mlp) across both models. This makes mlp 2 to be a good candidate for low-rank decomposition over the StarCoder model.

\textbf{Hardware Considerations:} On modern hardware accelerators like GPU and their corresponding software stack, matrix multiplication kernels are faster if their dimensions are divisible by a high factor of 2. So, we consider ranks at a reduction of approximately every 10\%, rounded off to the nearest multiple of 128 in our experiments.

\subsection{Performance of compressed models}\label{sec:compression_and_speedup_by_decomposition::subsec:Performance_of_compressed_models}

We consider the largest models of StarCoder and CodeGen family (16B) and perform low-rank decomposition on both with varying ranks. We consider decomposing layers that offers most parameter reduction (\S\ref{sec:compression_and_speedup_by_decomposition::subsec:Achieving_compression_and_inference_speedup}) with least increase in perplexity - mlp 2 for StarCoder and attention for CodeGen. We report the Pass@1 and Pass@10 scores over the Human Eval dataset \citep{humaneval_paper} using the code-eval GitHub repo \citep{anton_bacaj_code-eval_repository} in Table \ref{table:human_eval_lord_models}. We observe that StarCoder models can be low rank decomposed to 13.2B parameters (50\% rank reduction) with no drop in Pass@1 performance and upto 12.3B parameters (62.5\% rank reduction) with very little drop. CodeGen models shows similar trend in drop in Human Eval performance when measured in terms of rank reduction. However, in terms of parameter reduction count, while showing very little perplexity change with large reduction in rank (Fig. \ref{figure:codegen_layerwise_param_reduced_vs_perplexity}), shows much more drop in its HumanEval score when measured in terms of parameter count reduction due to a higher aspect ratio of the matrix being decomposed. It should be noted that for certain compressed models, the Pass@1 even slightly improves over the base model. Similar trend of slight improvements from compression across various metrics and benchmarks has been observed in the case of other compression attempts \citep{sparsegpt_elias_frantar_dan_alistarh, cerebras_sparse_blog}.

\subsection{Speedup from \methodnameshort{}}\label{sec:compression_and_speedup_by_decomposition::subsec:Speedup_from_LoRD}

We next consider accessing the inference speedup (forward pass) of the models over the standard cuBLAS floating point kernels. We consider the standard Huggingface implementation \citep{huggingface_paper} of Starcoder with pytorch backend \citep{pytorch_paper} utilizing standard cuBLAS kernels on A100 GPUs. \methodnameshort{} decomposed models were implemented by modifying just one line of code to replace an MLP with an extra linear layer \footnote{\scriptsize \texttt{nn.Linear(in, out) -> nn.Sequential(nn.Linear(in, rank), nn.Linear(rank, out))}}. We benchmark over 1024 tokens and 512 tokens sequence, averaged across 10 runs with warm up of 3 runs. We plot relative time taken and model size across reduction in rank in Figure 

\begin{wrapfigure}{r}{0.42\textwidth}
  \centering
  \includegraphics[width=0.4\textwidth]{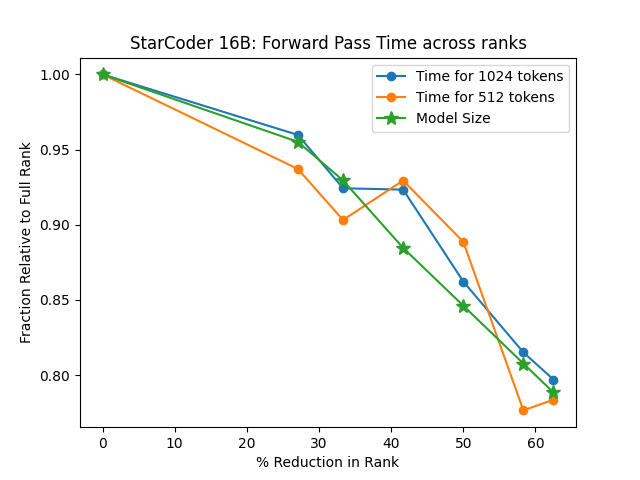}
  \caption{\label{figure:plot_relative_time_and_model_size} Time and Model size of StarCoder 16B across ranks.}
\end{wrapfigure}

\ref{figure:plot_relative_time_and_model_size}.

Inference speedups as high as 22.35\% are observed for decomposed models. The lines in the graph are generally downward sloping, Therefore reduction in rank beyond 25\% generally implies less inference time and reduction in model size. However, the underlying hardware (and pertaining software kernels) also significantly affect the speedup gains. We notice huge gains, whenever the rank is rounded off to a multiple of a very high power of 2 (like 4096 and 2560 at 33\% and 58\% rank reduction), despite very little reduction in model size. In contrast, for certain ranks which are multiples of a lesser power of 2 (like 3584 and 2304 at 41\% and 62\% rank reduction) are slower than those at slightly higher ranks. It is worth noting that affect of hardware inefficient matrix shape is less significant for longer tokens sequence of 1024 because the $O(n^2)$ attention overhead starts becoming more significant, especially in the absence of SoTA attention implementation techniques \citep{Self-attention_Does_Not_Need_O_n2__Memory_paper, flash_attention_paper, flash_attention_2_paper} as in the case of Huggingface's implementations.

\begin{table}
\scriptsize
\centering
\begin{tabular}{|c|c|c|c||c|c|c|c|}
\hline
\multicolumn{4}{|c||}{Starcoder 16B} & \multicolumn{4}{|c|}{CodeGen 16B Mono} \\
\hline
Model Type  & Rank & \multicolumn{2}{|c||}{HumanEval Score} & Model Type & Rank & \multicolumn{2}{|c|}{HumanEval Score} \\
\hline
\multicolumn{2}{|c|}{}                & Pass @ 1  & Pass @ 10 & \multicolumn{2}{|c|}{}               & Pass @ 1 & Pass @ 10 \\
\hline
Base Model                     & 6144 & 31.67     & 48.28     & Base Model                    & 6144 & 29.02    & 46.34    \\
\methodnameshort{}Coder 14.9B  & 4480 & 33.18     & 48.41     & \methodnameshort{}Coder 15.9B & 4480 & 29.08    & 46.95    \\
\methodnameshort{}Coder 14.5B  & 4096 & 31.69     & 45.12     & \methodnameshort{}Coder 15.6B & 4096 & 28.90    & 46.24    \\
\methodnameshort{}Coder 13.8B  & 3584 & 30.90     & 47.56     & \methodnameshort{}Coder 15.1B & 3584 & 28.54    & 45.73    \\
\methodnameshort{}Coder 13.2B  & 3072 & 31.57     & 45.36     & \methodnameshort{}Coder 14.7B & 3072 & 27.99    & 43.29    \\
\methodnameshort{}Coder 12.6B  & 2560 & 29.84     & 42.31     & \methodnameshort{}Coder 14.3B & 2560 & 27.32    & 45.12    \\
\methodnameshort{}Coder 12.3B  & 2304 & 29.22     & 40.12     & \methodnameshort{}Coder 14.1B & 2304 & 27.07    & 41.46    \\
\hline
\end{tabular}
\caption{\label{table:human_eval_lord_models} Human Eval Score of \methodnameshort{} across StarCoder and CodeGen.}
\end{table}

\section{Combining \methodnameshort{} with Quantization and LoRA}

\subsection{Quantization}\label{sec:combining_lord_with_quantization_and_lora::subsec:quantization}

While \methodnameshort{} enables compression at same precision level, we study whether the decomposed models can be further compressing through quantization. Table \ref{table:quantized_lordcoder_performance} shows the HumanEval pass@1 results for the different \methodnameshort{}Coder across 8 and 4 bit quantization levels, using the near lossless quantization technique of SpQR \citep{SpQR_tim_dettmers}. We observe that the \methodnameshort{} models can be combined with quantization for further compression, showing no performance drop for 8-bit and very little performance drop on 4-bit quantization for most models. Slight increase in HumanEval after quantization is also observed, similar to Pangu-Coder2 \citep{PanGu-Coder2}.

\begin{table}[h]
\scriptsize
\centering
\begin{tabular}{|l|l|l|l|}
\hline
Model                         & Pass@1@FP16 & Pass@1@8-bit & Pass@1@4-bit \\
\hline
\methodnameshort{}Coder 14.9B & 33.18       & 33.17        & 32.01        \\
\methodnameshort{}Coder 14.5B & 31.69       & 31.58        & 32.74        \\
\methodnameshort{}Coder 13.8B & 30.90       & 31.10        & 30.73        \\
\methodnameshort{}Coder 13.2B & 31.57       & 31.52        & 32.01        \\
\methodnameshort{}Coder 12.6B & 29.84       & 29.87        & 30.22        \\
\methodnameshort{}Coder 12.3B & 29.22       & 29.14        & 29.45        \\
\hline
\end{tabular}
\caption{\label{table:quantized_lordcoder_performance} Human Eval score of quantized LoRDCoder models.}

\end{table}

\subsection{Parameter Efficient tuning of \methodnameshort{} models}\label{sec:combining_lord_with_quantization_and_lora::subsec:Parameter_Efficient_tuning_of_lord}

\begin{wrapfigure}{r}{0.60\textwidth}
  \centering
  \includegraphics[width=0.60\textwidth]{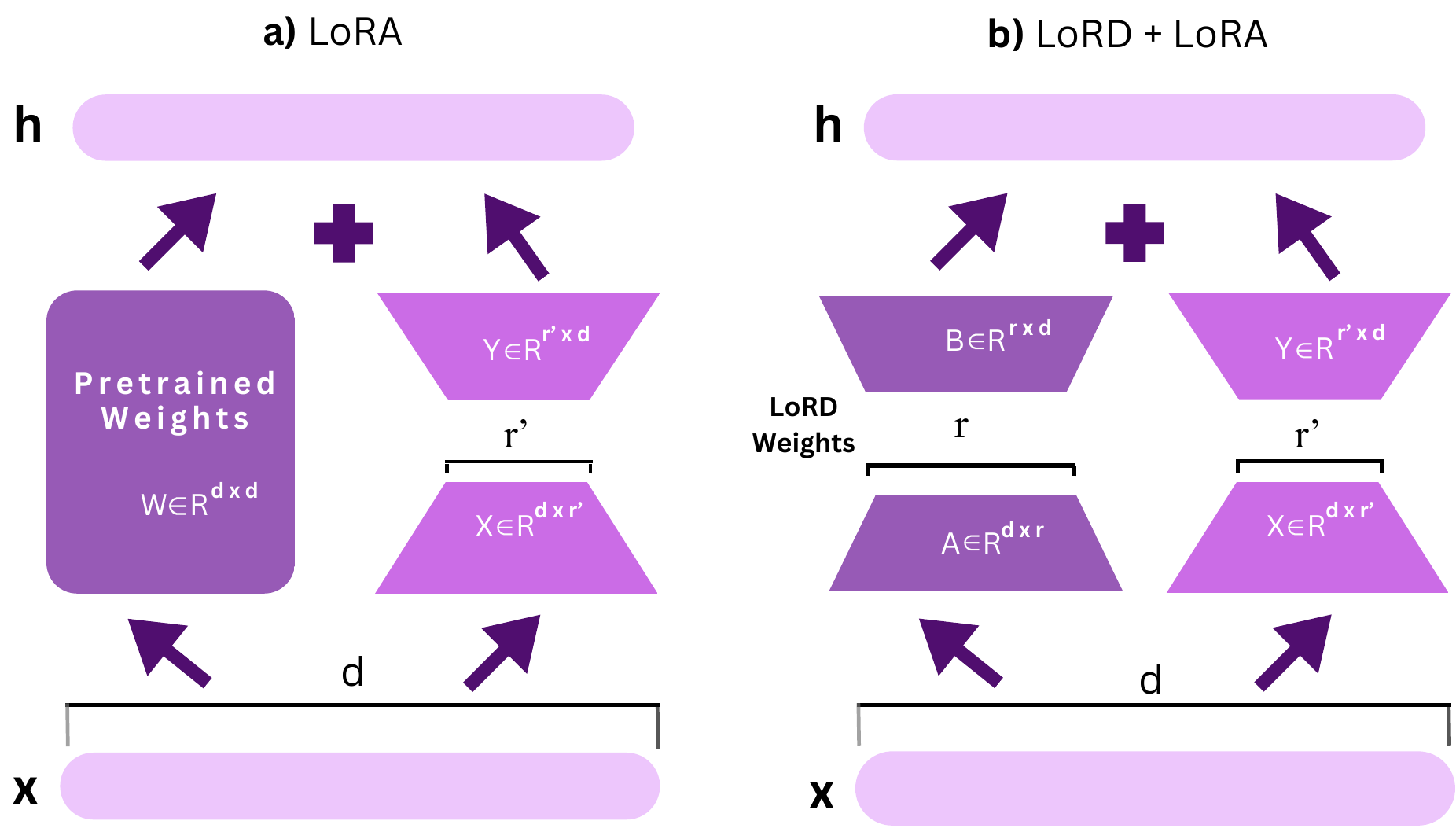}
  \caption{\label{figure:lora_vs_lord_lora} LoRA vs LoRD + LoRA.}
\end{wrapfigure}

We next test the potential for using \methodnameshort{} to further reduce the memory usage over existing parameter-efficient techniques. We consider the code instruction dataset \citep{sahil_glaive_https://huggingface.co/datasets/sahil2801/code_instructions_120k} and filter those examples that pertains to python programming language. We use QLoRA \citep{qlora_paper}, which is an even more memory efficient version of LoRA \citep{lora_paper} storing the weights in quantized format, for fine-tuning for 1 epoch. We compare results from fine-tuning two of the decomposed models \methodnameshort{}Coder 13.2B and \methodnameshort{}Coder 12.3B model to the StarCoder model. We observe a HumanEval pass@1 of 37.80 and 37.62 across \methodnameshort{}Coder 13.2B and \methodnameshort{}Coder 12.3B fine-tuning, competitive to the performance of 37.74 offered by StarCoder model.

% Add figure for LoRD vs Quantization vs LoRA vs LoRC

\section{Related Work}\label{section:related_work}

There is a growing interest in compressing pretrained Large Language Models. Several recent attempts have been dedicated to the quantization of weights of LLMs \citep{GPTQ_paper_elias_frantar_dan_alistarh, awq_paper, rptq_paper, lut_gemm_paper, finequant_paper, quip_2bit_quantization_paper, norm_tweaking_quantization_paper} with tricks such as outlier separation \citep{llm_int8_paper, the_case_for_4bit_precision_paper, spqr_paper, outlier_suppression_pushing_the_limit_paper, squeezellm_paper, owq_outlier_aware_weight_quantization_paper}. Some attempts also quantize the activations (intermediate representations) in addition to weights to speed up computation time \citep{omniquant_paper, smoothquant_paper}. The works in quantization that are closest to us is the Low-Rank Compensation (LoRC) Strategy \citep{ZeroQuant-V2_paper, ZeroQuant-FP_paper}, where the difference of the quantized matrix to the original matrix is approximated by a product of low-rank matrices. Our work decomposes the entire matrix for compression.

Pruning neural networks \cite{Pruning_and_quantization_for_deep_neural_network_acceleration:_A_survey}, unlike quantization, reduces the number of parameters in a model by removing unimportant weights or connections. Several techniques have been proposed to scale pruning methods for LLMs \citep{wanda_zico_kolter, sparsegpt_elias_frantar_dan_alistarh, LLM-Pruner:_On_the_Structural_Pruning_of_Large_Language_Models}. However, pruning as a means of compression is yet to become viable due to no speedups over sparse matrices without significant performance drop at extreme levels of sparsity or structured sparsity \citep{llm_compression_survey}. With low-rank decomposition, we propose an alternate method for reducing model parameters that offer speedup even at a little reduction in parameter count. Certain works have also attempted to \citep{Low_Rank_Prune-And-Factorize_for_Language_Model_Compression, LoSparse_paper} to split a dense matrix as a sum of low-rank matrices and a sparse matrix. However, these methods require retraining and have been shown to work only for Language Models of less than a billion parameters.

Low rank decomposition has been proposed for smaller language models like Bert or GPT2 before using SVD decomposition \citep{Compressing_Pre-trained_Language_Models_by_Matrix_Decomposition_yoav_goldeberg} and Kronecker decompositions \citep{kroneckerbert, Kronecker_Decomposition_for_GPT_Compression}. \cite{Language_model_compression_with_weighted_lowrank_factorization} modified SVD to be data aware based on approximate second-order gradient information. A better weighted SVD was proposed by \citep{Numerical_Optimizations_for_Weighted_Low-rank_Estimation_on_Language_Model}.
\cite{DRONE_Dataaware_Lowrank_Compression_for_Large_NLP_Models} proposed a data aware decomposition method with a provably optimal closed-form solution, utilizing a large amount of data points over specific tasks to decompose. Several recent works \citep{Compressing_transformers_features_are_lowrank_but_weights_are_not, rank_diminishing_in_deep_neural_networks_michael_jordan} have shown that while the weight matrix of neural networks is not inherently low-rank, the intermediate representations are, thus propose to decompose based on representations. All these works have focused on small language models and require re-training. We proposed low-rank decomposition for compressing neural networks without the need for retraining. The factorization has also been used just for the embedding layers \citep{embedding_matrix_factorization_Adaptive_Input_Representations_for_Neural_Language_Modeling, embedding_matrix_factorization_ALBERT:_A_Lite_BERT_for_Self-supervised_Learning_of_Language_Representations}, as they are good candidates due to their very low aspect ratio of 0.015, where a reduction of rank by even 5\% would lead to reduction in number of parameters after decomposition.

There is also a growing interest in fine-tuning large language models \cite{alpaca_paper, vicuna_paper, camel_paper, dromedary_paper}. With the large memory requirements for fine-tuning full parameters of the LLM, the more parameter-efficient fine-tuning methods like LoRA \citep{lora_paper} are getting widely adopted. These methods freeze the original LLM weights, and attach two low-rank matrices or adapters, in a skip-connection \citep{resnet_paper} to the linear layers of the model. These parameter-efficient fine-tuning approaches have seen improvements in lower activation memory \citep{LORA-FA:_MEMORY-EFFICIENT} or by keeping non-trainable model weights at 4-bit precision \citep{qlora_paper}. Our work, while focused on compression through low-rank decomposition, can also enable more efficient fine-tuning, especially in conjunction with existing methods.

\section{Conclusion}

We studied the compression of monolingual code generation models through a novel one-shot compression paradigm of low-rank decomposition. We analyse the change in perplexity with change in rank across the model families of StarCoder and CodeGen as well as their individual layers and observe that the rank of these models can be reduced by upto 39.58\% with less than 1\% change in perplexity. We then proposed considerations for one-shot compressing these models through \methodname{} in under 10 minutes. Consequently, we compress StarCoder 16B to 13.2B with no drop in HumanEval pass@1 and very little drop in HumanEval pass@1 to 12.3B parameters. With a minimal change in code over huggingface's default inference code of just one line, we gain speedups of up to 22.35\%. The \methodnameshort{} models are also compatible with near lossless quantization techniques of SpQR, which offers gains of quantization based compression in addition to ones from decomposition. The \methodnameshort{} models also reduce memory requirements by as much as 21.2\% over vanilla QLoRA fine-tuning.

\section{Broader Impact and Future Work}

Our work on \methodnameshort{}, compresses code LLMs which enables them to run on smaller GPUs including as consumer grade GPUs. This is especially of pressing importance for the next few years when the shortage of GPU supply is relative to the increasing demand in today's market. Moreover, faster inference helps reduce the GPU cycles, enabling lower running costs and lower power consumption for LLM inference. Our work helps reduce the carbon emissions incurred and moves towards a greener NLP. Through compression, our work also promotes inference at the edge, and therefore opening room for applications involving strict privacy requirements. Lower latency will also help improve the User Experience in applications like CoPilots where lag between suggestions can impact developer's productivity. Several of these benefits of \methodnameshort{} such as lower cost and energy consumption are also applicable for fine-tuning use cases of LLMs.

Our work opens up a new paradigm for compression via Low Rank Decomposition over Large Language Models in a single shot without the need for retraining. Since, \methodnameshort{} models can leverage existing floating point kernels across BLAS and cuBLAS, in contrast to quantization, these are much easier to implement and reap inference benefits. Our study on hardware considerations for speedup also opens up the potential for tuning the rank of decomposed models to fit best on the target hardware and the accompanying GEMM kernels. While our study is limited to monolingual code LLMs, the low rank decomposition technique is general and not specific to code domain. Thus exploring its applicability to more general purpose models like LLaMa is a promising direction for the compression of transformer LLMs beyond quantization. Another interesting unexplored question is whether the LoRA or QLoRA modules fine-tuned on original models, can be plugged in as-is for the \methodnameshort{} models without any performance drop.



\bibliography{iclr2021_conference}

\begin{thebibliography}{72}
\providecommand{\natexlab}[1]{#1}
\providecommand{\url}[1]{\texttt{#1}}
\expandafter\ifx\csname urlstyle\endcsname\relax
  \providecommand{\doi}[1]{doi: #1}\else
  \providecommand{\doi}{doi: \begingroup \urlstyle{rm}\Url}\fi

\bibitem[Agarwal et~al.(2023)Agarwal, Vieillard, Stanczyk, Ramos, Geist, and
  Bachem]{gkd_distillation_deepmind}
Rishabh Agarwal, Nino Vieillard, Piotr Stanczyk, Sabela Ramos, Matthieu Geist,
  and Olivier Bachem.
\newblock Gkd: Generalized knowledge distillation for auto-regressive sequence
  models, 2023.

\bibitem[Bacaj(2023)]{anton_bacaj_code-eval_repository}
Anton Bacaj.
\newblock code-eval.
\newblock \url{https://github.com/abacaj/code-eval}, July 2023.

\bibitem[Baevski \& Auli(2019)Baevski and
  Auli]{embedding_matrix_factorization_Adaptive_Input_Representations_for_Neural_Language_Modeling}
Alexei Baevski and Michael Auli.
\newblock Adaptive input representations for neural language modeling.
\newblock In \emph{International Conference on Learning Representations}, 2019.
\newblock URL \url{https://openreview.net/forum?id=ByxZX20qFQ}.

\bibitem[Ben~Noach \& Goldberg(2020)Ben~Noach and
  Goldberg]{Compressing_Pre-trained_Language_Models_by_Matrix_Decomposition_yoav_goldeberg}
Matan Ben~Noach and Yoav Goldberg.
\newblock Compressing pre-trained language models by matrix decomposition.
\newblock In \emph{Proceedings of the 1st Conference of the Asia-Pacific
  Chapter of the Association for Computational Linguistics and the 10th
  International Joint Conference on Natural Language Processing}, pp.\
  884--889, Suzhou, China, December 2020. Association for Computational
  Linguistics.
\newblock URL \url{https://aclanthology.org/2020.aacl-main.88}.

\bibitem[Bigcode(2022)]{the_stack_smol_huggignface}
Project Bigcode.
\newblock The stack smol, 2022.
\newblock URL \url{https://huggingface.co/datasets/bigcode/the-stack-smol}.

\bibitem[Blackford et~al.(2002)Blackford, Petitet, Pozo, Remington, Whaley,
  Demmel, Dongarra, Duff, Hammarling, Henry, et~al.]{blas_citation}
L~Susan Blackford, Antoine Petitet, Roldan Pozo, Karin Remington, R~Clint
  Whaley, James Demmel, Jack Dongarra, Iain Duff, Sven Hammarling, Greg Henry,
  et~al.
\newblock An updated set of basic linear algebra subprograms (blas).
\newblock \emph{ACM Transactions on Mathematical Software}, 28\penalty0
  (2):\penalty0 135--151, 2002.

\bibitem[Cerebras(2022)]{cerebras_sparse_blog}
Team Cerebras.
\newblock Creating sparse gpt-3 models with iterative pruning, 11 2022.
\newblock URL
  \url{https://www.cerebras.net/blog/creating-sparse-gpt-3-models-with-iterative-pruning}.

\bibitem[Chaudhary(2023)]{sahil_glaive_https://huggingface.co/datasets/sahil2801/code_instructions_120k}
Sahil Chaudhary.
\newblock Code instructions dataset.
\newblock
  \url{https://huggingface.co/datasets/sahil2801/code_instructions_120k}, Jun
  2023.

\bibitem[Chee et~al.(2023)Chee, Cai, Kuleshov, and
  Sa]{quip_2bit_quantization_paper}
Jerry Chee, Yaohui Cai, Volodymyr Kuleshov, and Christopher~De Sa.
\newblock Quip: 2-bit quantization of large language models with guarantees,
  2023.

\bibitem[Chen et~al.(2021{\natexlab{a}})Chen, Tworek, Jun, Yuan,
  de~Oliveira~Pinto, Kaplan, Edwards, Burda, Joseph, Brockman, Ray, Puri,
  Krueger, Petrov, Khlaaf, Sastry, Mishkin, Chan, Gray, Ryder, Pavlov, Power,
  Kaiser, Bavarian, Winter, Tillet, Such, Cummings, Plappert, Chantzis, Barnes,
  Herbert-Voss, Guss, Nichol, Paino, Tezak, Tang, Babuschkin, Balaji, Jain,
  Saunders, Hesse, Carr, Leike, Achiam, Misra, Morikawa, Radford, Knight,
  Brundage, Murati, Mayer, Welinder, McGrew, Amodei, McCandlish, Sutskever, and
  Zaremba]{humaneval_paper}
Mark Chen, Jerry Tworek, Heewoo Jun, Qiming Yuan, Henrique~Ponde
  de~Oliveira~Pinto, Jared Kaplan, Harri Edwards, Yuri Burda, Nicholas Joseph,
  Greg Brockman, Alex Ray, Raul Puri, Gretchen Krueger, Michael Petrov, Heidy
  Khlaaf, Girish Sastry, Pamela Mishkin, Brooke Chan, Scott Gray, Nick Ryder,
  Mikhail Pavlov, Alethea Power, Lukasz Kaiser, Mohammad Bavarian, Clemens
  Winter, Philippe Tillet, Felipe~Petroski Such, Dave Cummings, Matthias
  Plappert, Fotios Chantzis, Elizabeth Barnes, Ariel Herbert-Voss,
  William~Hebgen Guss, Alex Nichol, Alex Paino, Nikolas Tezak, Jie Tang, Igor
  Babuschkin, Suchir Balaji, Shantanu Jain, William Saunders, Christopher
  Hesse, Andrew~N. Carr, Jan Leike, Josh Achiam, Vedant Misra, Evan Morikawa,
  Alec Radford, Matthew Knight, Miles Brundage, Mira Murati, Katie Mayer, Peter
  Welinder, Bob McGrew, Dario Amodei, Sam McCandlish, Ilya Sutskever, and
  Wojciech Zaremba.
\newblock Evaluating large language models trained on code, 2021{\natexlab{a}}.

\bibitem[Chen et~al.(2021{\natexlab{b}})Chen, Yu, Dhillon, and
  Hsieh]{DRONE_Dataaware_Lowrank_Compression_for_Large_NLP_Models}
Patrick Chen, Hsiang-Fu Yu, Inderjit Dhillon, and Cho-Jui Hsieh.
\newblock Drone: Data-aware low-rank compression for large nlp models.
\newblock In M.~Ranzato, A.~Beygelzimer, Y.~Dauphin, P.S. Liang, and J.~Wortman
  Vaughan (eds.), \emph{Advances in Neural Information Processing Systems},
  volume~34, pp.\  29321--29334. Curran Associates, Inc., 2021{\natexlab{b}}.
\newblock URL
  \url{https://proceedings.neurips.cc/paper_files/paper/2021/file/f56de5ef149cf0aedcc8f4797031e229-Paper.pdf}.

\bibitem[Chiang et~al.(2023)Chiang, Li, Lin, Sheng, Wu, Zhang, Zheng, Zhuang,
  Zhuang, Gonzalez, Stoica, and Xing]{vicuna_paper}
Wei-Lin Chiang, Zhuohan Li, Zi~Lin, Ying Sheng, Zhanghao Wu, Hao Zhang, Lianmin
  Zheng, Siyuan Zhuang, Yonghao Zhuang, Joseph~E. Gonzalez, Ion Stoica, and
  Eric~P. Xing.
\newblock Vicuna: An open-source chatbot impressing gpt-4 with 90\%* chatgpt
  quality, March 2023.
\newblock URL \url{https://lmsys.org/blog/2023-03-30-vicuna/}.

\bibitem[Dao(2023)]{flash_attention_2_paper}
Tri Dao.
\newblock Flashattention-2: Faster attention with better parallelism and work
  partitioning, 2023.

\bibitem[Dao et~al.(2022)Dao, Fu, Ermon, Rudra, and Re]{flash_attention_paper}
Tri Dao, Daniel~Y Fu, Stefano Ermon, Atri Rudra, and Christopher Re.
\newblock Flashattention: Fast and memory-efficient exact attention with
  {IO}-awareness.
\newblock In Alice~H. Oh, Alekh Agarwal, Danielle Belgrave, and Kyunghyun Cho
  (eds.), \emph{Advances in Neural Information Processing Systems}, 2022.
\newblock URL \url{https://openreview.net/forum?id=H4DqfPSibmx}.

\bibitem[Dettmers \& Zettlemoyer(2022)Dettmers and
  Zettlemoyer]{the_case_for_4bit_precision_paper}
Tim Dettmers and Luke Zettlemoyer.
\newblock The case for 4-bit precision: k-bit inference scaling laws, 2022.

\bibitem[Dettmers et~al.(2022)Dettmers, Lewis, Belkada, and
  Zettlemoyer]{llm_int8_paper}
Tim Dettmers, Mike Lewis, Younes Belkada, and Luke Zettlemoyer.
\newblock {GPT}3.int8(): 8-bit matrix multiplication for transformers at scale.
\newblock In Alice~H. Oh, Alekh Agarwal, Danielle Belgrave, and Kyunghyun Cho
  (eds.), \emph{Advances in Neural Information Processing Systems}, 2022.
\newblock URL \url{https://openreview.net/forum?id=dXiGWqBoxaD}.

\bibitem[Dettmers et~al.(2023{\natexlab{a}})Dettmers, Pagnoni, Holtzman, and
  Zettlemoyer]{qlora_paper}
Tim Dettmers, Artidoro Pagnoni, Ari Holtzman, and Luke Zettlemoyer.
\newblock Qlora: Efficient finetuning of quantized llms, 2023{\natexlab{a}}.

\bibitem[Dettmers et~al.(2023{\natexlab{b}})Dettmers, Svirschevski, Egiazarian,
  Kuznedelev, Frantar, Ashkboos, Borzunov, Hoefler, and
  Alistarh]{SpQR_tim_dettmers}
Tim Dettmers, Ruslan Svirschevski, Vage Egiazarian, Denis Kuznedelev, Elias
  Frantar, Saleh Ashkboos, Alexander Borzunov, Torsten Hoefler, and Dan
  Alistarh.
\newblock Spqr: A sparse-quantized representation for near-lossless llm weight
  compression, 2023{\natexlab{b}}.

\bibitem[Dettmers et~al.(2023{\natexlab{c}})Dettmers, Svirschevski, Egiazarian,
  Kuznedelev, Frantar, Ashkboos, Borzunov, Hoefler, and Alistarh]{spqr_paper}
Tim Dettmers, Ruslan Svirschevski, Vage Egiazarian, Denis Kuznedelev, Elias
  Frantar, Saleh Ashkboos, Alexander Borzunov, Torsten Hoefler, and Dan
  Alistarh.
\newblock Spqr: A sparse-quantized representation for near-lossless llm weight
  compression, 2023{\natexlab{c}}.

\bibitem[Devlin et~al.(2019)Devlin, Chang, Lee, and
  Toutanova]{bert_paper_devlin_et_al}
Jacob Devlin, Ming-Wei Chang, Kenton Lee, and Kristina Toutanova.
\newblock {BERT}: Pre-training of deep bidirectional transformers for language
  understanding.
\newblock In \emph{Proceedings of the 2019 Conference of the North {A}merican
  Chapter of the Association for Computational Linguistics: Human Language
  Technologies, Volume 1 (Long and Short Papers)}, pp.\  4171--4186,
  Minneapolis, Minnesota, June 2019. Association for Computational Linguistics.
\newblock \doi{10.18653/v1/N19-1423}.
\newblock URL \url{https://aclanthology.org/N19-1423}.

\bibitem[Edalati et~al.(2022)Edalati, Tahaei, Rashid, Nia, Clark, and
  Rezagholizadeh]{Kronecker_Decomposition_for_GPT_Compression}
Ali Edalati, Marzieh Tahaei, Ahmad Rashid, Vahid Nia, James Clark, and Mehdi
  Rezagholizadeh.
\newblock Kronecker decomposition for {GPT} compression.
\newblock In \emph{Proceedings of the 60th Annual Meeting of the Association
  for Computational Linguistics (Volume 2: Short Papers)}, pp.\  219--226,
  Dublin, Ireland, May 2022. Association for Computational Linguistics.
\newblock \doi{10.18653/v1/2022.acl-short.24}.
\newblock URL \url{https://aclanthology.org/2022.acl-short.24}.

\bibitem[Feng et~al.(2022)Feng, Zheng, Huang, Zhao, Jordan, and
  Zha]{rank_diminishing_in_deep_neural_networks_michael_jordan}
Ruili Feng, Kecheng Zheng, Yukun Huang, Deli Zhao, Michael Jordan, and
  Zheng-Jun Zha.
\newblock Rank diminishing in deep neural networks.
\newblock In Alice~H. Oh, Alekh Agarwal, Danielle Belgrave, and Kyunghyun Cho
  (eds.), \emph{Advances in Neural Information Processing Systems}, 2022.
\newblock URL \url{https://openreview.net/forum?id=tIqzLFf3kk}.

\bibitem[Frantar \& Alistarh(2023)Frantar and
  Alistarh]{sparsegpt_elias_frantar_dan_alistarh}
Elias Frantar and Dan Alistarh.
\newblock Sparsegpt: Massive language models can be accurately pruned in
  one-shot, 2023.

\bibitem[Frantar et~al.(2023)Frantar, Ashkboos, Hoefler, and
  Alistarh]{GPTQ_paper_elias_frantar_dan_alistarh}
Elias Frantar, Saleh Ashkboos, Torsten Hoefler, and Dan Alistarh.
\newblock {OPTQ}: Accurate quantization for generative pre-trained
  transformers.
\newblock In \emph{The Eleventh International Conference on Learning
  Representations}, 2023.
\newblock URL \url{https://openreview.net/forum?id=tcbBPnfwxS}.

\bibitem[Gu et~al.(2023)Gu, Dong, Wei, and
  Huang]{minillm_distillation_microsoft}
Yuxian Gu, Li~Dong, Furu Wei, and Minlie Huang.
\newblock Knowledge distillation of large language models, 2023.

\bibitem[He et~al.(2016)He, Zhang, Ren, and Sun]{resnet_paper}
Kaiming He, Xiangyu Zhang, Shaoqing Ren, and Jian Sun.
\newblock Deep residual learning for image recognition.
\newblock In \emph{2016 IEEE Conference on Computer Vision and Pattern
  Recognition (CVPR)}, pp.\  770--778, 2016.
\newblock \doi{10.1109/CVPR.2016.90}.

\bibitem[Hinton et~al.(2015)Hinton, Vinyals, and
  Dean]{hinton_knowledge_distillation}
Geoffrey Hinton, Oriol Vinyals, and Jeff Dean.
\newblock Distilling the knowledge in a neural network, 2015.

\bibitem[Hsu et~al.(2022)Hsu, Hua, Chang, Lou, Shen, and
  Jin]{Language_model_compression_with_weighted_lowrank_factorization}
Yen-Chang Hsu, Ting Hua, Sungen Chang, Qian Lou, Yilin Shen, and Hongxia Jin.
\newblock Language model compression with weighted low-rank factorization.
\newblock In \emph{International Conference on Learning Representations}, 2022.
\newblock URL \url{https://openreview.net/forum?id=uPv9Y3gmAI5}.

\bibitem[Hu et~al.(2022)Hu, yelong shen, Wallis, Allen-Zhu, Li, Wang, Wang, and
  Chen]{lora_paper}
Edward~J Hu, yelong shen, Phillip Wallis, Zeyuan Allen-Zhu, Yuanzhi Li, Shean
  Wang, Lu~Wang, and Weizhu Chen.
\newblock Lo{RA}: Low-rank adaptation of large language models.
\newblock In \emph{International Conference on Learning Representations}, 2022.
\newblock URL \url{https://openreview.net/forum?id=nZeVKeeFYf9}.

\bibitem[Hua et~al.(2022)Hua, Hsu, Wang, Lou, Shen, and
  Jin]{Numerical_Optimizations_for_Weighted_Low-rank_Estimation_on_Language_Model}
Ting Hua, Yen-Chang Hsu, Felicity Wang, Qian Lou, Yilin Shen, and Hongxia Jin.
\newblock Numerical optimizations for weighted low-rank estimation on language
  models.
\newblock In \emph{Proceedings of the 2022 Conference on Empirical Methods in
  Natural Language Processing}, pp.\  1404--1416, Abu Dhabi, United Arab
  Emirates, December 2022. Association for Computational Linguistics.
\newblock \doi{10.18653/v1/2022.emnlp-main.91}.
\newblock URL \url{https://aclanthology.org/2022.emnlp-main.91}.

\bibitem[Jung et~al.(2023)Jung, West, Jiang, Brahman, Lu, Fisher, Sorensen, and
  Choi]{impossible_distillation_yejin_choi}
Jaehun Jung, Peter West, Liwei Jiang, Faeze Brahman, Ximing Lu, Jillian Fisher,
  Taylor Sorensen, and Yejin Choi.
\newblock Impossible distillation: from low-quality model to high-quality
  dataset \& model for summarization and paraphrasing, 2023.

\bibitem[Kim et~al.(2023{\natexlab{a}})Kim, Hooper, Gholami, Dong, Li, Shen,
  Mahoney, and Keutzer]{squeezellm_paper}
Sehoon Kim, Coleman Hooper, Amir Gholami, Zhen Dong, Xiuyu Li, Sheng Shen,
  Michael~W. Mahoney, and Kurt Keutzer.
\newblock Squeezellm: Dense-and-sparse quantization, 2023{\natexlab{a}}.

\bibitem[Kim et~al.(2023{\natexlab{b}})Kim, Henry, Fahim, and
  Awadalla]{finequant_paper}
Young~Jin Kim, Rawn Henry, Raffy Fahim, and Hany~Hassan Awadalla.
\newblock Finequant: Unlocking efficiency with fine-grained weight-only
  quantization for llms, 2023{\natexlab{b}}.

\bibitem[Kocetkov et~al.(2022)Kocetkov, Li, Allal, Li, Mou, Ferrandis, Jernite,
  Mitchell, Hughes, Wolf, Bahdanau, von Werra, and de~Vries]{the_stack_paper}
Denis Kocetkov, Raymond Li, Loubna~Ben Allal, Jia Li, Chenghao Mou,
  Carlos~Muñoz Ferrandis, Yacine Jernite, Margaret Mitchell, Sean Hughes,
  Thomas Wolf, Dzmitry Bahdanau, Leandro von Werra, and Harm de~Vries.
\newblock The stack: 3 tb of permissively licensed source code, 2022.

\bibitem[Lan et~al.(2020)Lan, Chen, Goodman, Gimpel, Sharma, and
  Soricut]{embedding_matrix_factorization_ALBERT:_A_Lite_BERT_for_Self-supervised_Learning_of_Language_Representations}
Zhenzhong Lan, Mingda Chen, Sebastian Goodman, Kevin Gimpel, Piyush Sharma, and
  Radu Soricut.
\newblock Albert: A lite bert for self-supervised learning of language
  representations.
\newblock In \emph{International Conference on Learning Representations}, 2020.
\newblock URL \url{https://openreview.net/forum?id=H1eA7AEtvS}.

\bibitem[Lee et~al.(2023)Lee, Jin, Kim, Kim, and
  Park]{owq_outlier_aware_weight_quantization_paper}
Changhun Lee, Jungyu Jin, Taesu Kim, Hyungjun Kim, and Eunhyeok Park.
\newblock Owq: Lessons learned from activation outliers for weight quantization
  in large language models, 2023.

\bibitem[Li et~al.(2023{\natexlab{a}})Li, Li, Zhang, and
  Chu]{norm_tweaking_quantization_paper}
Liang Li, Qingyuan Li, Bo~Zhang, and Xiangxiang Chu.
\newblock Norm tweaking: High-performance low-bit quantization of large
  language models, 2023{\natexlab{a}}.

\bibitem[Li et~al.(2023{\natexlab{b}})Li, Yu, Zhang, Liang, He, Chen, and
  Zhao]{LoSparse_paper}
Yixiao Li, Yifan Yu, Qingru Zhang, Chen Liang, Pengcheng He, Weizhu Chen, and
  Tuo Zhao.
\newblock Losparse: Structured compression of large language models based on
  low-rank and sparse approximation, 2023{\natexlab{b}}.

\bibitem[Li et~al.(2020)Li, Wallace, Shen, Lin, Keutzer, Klein, and
  Gonzalez]{train_large_then_compress}
Zhuohan Li, Eric Wallace, Sheng Shen, Kevin Lin, Kurt Keutzer, Dan Klein, and
  Joseph~E. Gonzalez.
\newblock Train large, then compress: Rethinking model size for efficient
  training and inference of transformers.
\newblock In \emph{Proceedings of the 37th International Conference on Machine
  Learning}, ICML'20. JMLR.org, 2020.

\bibitem[Liang et~al.(2021)Liang, Glossner, Wang, Shi, and
  Zhang]{Pruning_and_quantization_for_deep_neural_network_acceleration:_A_survey}
Tailin Liang, John Glossner, Lei Wang, Shaobo Shi, and Xiaotong Zhang.
\newblock Pruning and quantization for deep neural network acceleration: A
  survey.
\newblock \emph{Neurocomputing}, 461:\penalty0 370--403, 2021.
\newblock ISSN 0925-2312.
\newblock \doi{https://doi.org/10.1016/j.neucom.2021.07.045}.
\newblock URL
  \url{https://www.sciencedirect.com/science/article/pii/S0925231221010894}.

\bibitem[Lin et~al.(2023)Lin, Tang, Tang, Yang, Dang, and Han]{awq_paper}
Ji~Lin, Jiaming Tang, Haotian Tang, Shang Yang, Xingyu Dang, and Song Han.
\newblock Awq: Activation-aware weight quantization for llm compression and
  acceleration, 2023.

\bibitem[Ma et~al.(2023)Ma, Fang, and
  Wang]{LLM-Pruner:_On_the_Structural_Pruning_of_Large_Language_Models}
Xinyin Ma, Gongfan Fang, and Xinchao Wang.
\newblock Llm-pruner: On the structural pruning of large language models, 2023.

\bibitem[NVIDIA(2007)]{cuda}
Corporation NVIDIA.
\newblock Compute unified device architecture (cuda).
\newblock Website, 2007.
\newblock URL \url{https://developer.nvidia.com/cuda-toolkit}.
\newblock Accessed: 2023-09-17.

\bibitem[Park et~al.(2022)Park, Park, Kim, Lee, Kim, Kwon, Kwon, Kim, Lee, and
  Lee]{lut_gemm_paper}
Gunho Park, Baeseong Park, Minsub Kim, Sungjae Lee, Jeonghoon Kim, Beomseok
  Kwon, Se~Jung Kwon, Byeongwook Kim, Youngjoo Lee, and Dongsoo Lee.
\newblock Lut-gemm: Quantized matrix multiplication based on luts for efficient
  inference in large-scale generative language models, 2022.

\bibitem[Paszke et~al.(2019)Paszke, Gross, Massa, Lerer, Bradbury, Chanan,
  Killeen, Lin, Gimelshein, Antiga, Desmaison, K\"{o}pf, Yang, DeVito, Raison,
  Tejani, Chilamkurthy, Steiner, Fang, Bai, and Chintala]{pytorch_paper}
Adam Paszke, Sam Gross, Francisco Massa, Adam Lerer, James Bradbury, Gregory
  Chanan, Trevor Killeen, Zeming Lin, Natalia Gimelshein, Luca Antiga, Alban
  Desmaison, Andreas K\"{o}pf, Edward Yang, Zach DeVito, Martin Raison, Alykhan
  Tejani, Sasank Chilamkurthy, Benoit Steiner, Lu~Fang, Junjie Bai, and Soumith
  Chintala.
\newblock \emph{PyTorch: An Imperative Style, High-Performance Deep Learning
  Library}, chapter~., pp.\ ~.
\newblock Curran Associates Inc., Red Hook, NY, USA, 2019.

\bibitem[Penedo et~al.(2023)Penedo, Malartic, Hesslow, Cojocaru, Cappelli,
  Alobeidli, Pannier, Almazrouei, and Launay]{falcon_paper}
Guilherme Penedo, Quentin Malartic, Daniel Hesslow, Ruxandra Cojocaru,
  Alessandro Cappelli, Hamza Alobeidli, Baptiste Pannier, Ebtesam Almazrouei,
  and Julien Launay.
\newblock The refinedweb dataset for falcon llm: Outperforming curated corpora
  with web data, and web data only, 2023.

\bibitem[Peng et~al.(2023)Peng, Kalliamvakou, Cihon, and
  Demirer]{copilot_productivity_gains}
Sida Peng, Eirini Kalliamvakou, Peter Cihon, and Mert Demirer.
\newblock The impact of ai on developer productivity: Evidence from github
  copilot, 2023.

\bibitem[Rabe \& Staats(2021)Rabe and
  Staats]{Self-attention_Does_Not_Need_O_n2__Memory_paper}
Markus~N. Rabe and Charles Staats.
\newblock Self-attention does not need $o(n^2)$ memory, 2021.

\bibitem[Radford et~al.(2019)Radford, Wu, Child, Luan, Amodei, and
  Sutskever]{gpt2_paper}
Alec Radford, Jeff Wu, Rewon Child, David Luan, Dario Amodei, and Ilya
  Sutskever.
\newblock Language models are unsupervised multitask learners.
\newblock \emph{OpenAI Blog}, 1\penalty0 (8), 2019.

\bibitem[Ren \& Zhu(2023)Ren and
  Zhu]{Low_Rank_Prune-And-Factorize_for_Language_Model_Compression}
Siyu Ren and Kenny~Q. Zhu.
\newblock Low-rank prune-and-factorize for language model compression, 2023.

\bibitem[Rozière et~al.(2023)Rozière, Gehring, Gloeckle, Sootla, Gat, Tan,
  Adi, Liu, Remez, Rapin, Kozhevnikov, Evtimov, Bitton, Bhatt, Ferrer,
  Grattafiori, Xiong, Défossez, Copet, Azhar, Touvron, Martin, Usunier,
  Scialom, and Synnaeve]{codellama_paper}
Baptiste Rozière, Jonas Gehring, Fabian Gloeckle, Sten Sootla, Itai Gat,
  Xiaoqing~Ellen Tan, Yossi Adi, Jingyu Liu, Tal Remez, Jérémy Rapin, Artyom
  Kozhevnikov, Ivan Evtimov, Joanna Bitton, Manish Bhatt, Cristian~Canton
  Ferrer, Aaron Grattafiori, Wenhan Xiong, Alexandre Défossez, Jade Copet,
  Faisal Azhar, Hugo Touvron, Louis Martin, Nicolas Usunier, Thomas Scialom,
  and Gabriel Synnaeve.
\newblock Code llama: Open foundation models for code, 2023.

\bibitem[Sanh et~al.(2019)Sanh, Debut, Chaumond, and Wolf]{distilbert_paper}
Victor Sanh, Lysandre Debut, Julien Chaumond, and Thomas Wolf.
\newblock Distilbert, a distilled version of bert: smaller, faster, cheaper and
  lighter, 2019.

\bibitem[Shao et~al.(2023)Shao, Chen, Zhang, Xu, Zhao, Li, Zhang, Gao, Qiao,
  and Luo]{omniquant_paper}
Wenqi Shao, Mengzhao Chen, Zhaoyang Zhang, Peng Xu, Lirui Zhao, Zhiqian Li,
  Kaipeng Zhang, Peng Gao, Yu~Qiao, and Ping Luo.
\newblock Omniquant: Omnidirectionally calibrated quantization for large
  language models, 2023.

\bibitem[Shazeer(2019)]{multi_query_attention_paper}
Noam Shazeer.
\newblock Fast transformer decoding: One write-head is all you need, 2019.

\bibitem[Shazeer(2020)]{swiglu_paper}
Noam Shazeer.
\newblock Glu variants improve transformer, 2020.

\bibitem[Shen et~al.(2023)Shen, Zhang, Chen, Zan, Geng, Fu, Zeng, Yu, Ji, Zhao,
  Guo, and Wang]{PanGu-Coder2}
Bo~Shen, Jiaxin Zhang, Taihong Chen, Daoguang Zan, Bing Geng, An~Fu, Muhan
  Zeng, Ailun Yu, Jichuan Ji, Jingyang Zhao, Yuenan Guo, and Qianxiang Wang.
\newblock Pangu-coder2: Boosting large language models for code with ranking
  feedback, 2023.

\bibitem[Sun et~al.(2023{\natexlab{a}})Sun, Liu, Bair, and
  Kolter]{wanda_zico_kolter}
Mingjie Sun, Zhuang Liu, Anna Bair, and J.~Zico Kolter.
\newblock A simple and effective pruning approach for large language models,
  2023{\natexlab{a}}.

\bibitem[Sun et~al.(2023{\natexlab{b}})Sun, Shen, Zhou, Zhang, Chen, Cox, Yang,
  and Gan]{dromedary_paper}
Zhiqing Sun, Yikang Shen, Qinhong Zhou, Hongxin Zhang, Zhenfang Chen, David
  Cox, Yiming Yang, and Chuang Gan.
\newblock Principle-driven self-alignment of language models from scratch with
  minimal human supervision, 2023{\natexlab{b}}.

\bibitem[Tahaei et~al.(2022)Tahaei, Charlaix, Nia, Ghodsi, and
  Rezagholizadeh]{kroneckerbert}
Marzieh Tahaei, Ella Charlaix, Vahid Nia, Ali Ghodsi, and Mehdi Rezagholizadeh.
\newblock {K}ronecker{BERT}: Significant compression of pre-trained language
  models through kronecker decomposition and knowledge distillation.
\newblock In \emph{Proceedings of the 2022 Conference of the North American
  Chapter of the Association for Computational Linguistics: Human Language
  Technologies}, pp.\  2116--2127, Seattle, United States, July 2022.
  Association for Computational Linguistics.
\newblock \doi{10.18653/v1/2022.naacl-main.154}.
\newblock URL \url{https://aclanthology.org/2022.naacl-main.154}.

\bibitem[Taori et~al.(2023)Taori, Gulrajani, Zhang, Dubois, Li, Guestrin,
  Liang, and Hashimoto]{alpaca_paper}
Rohan Taori, Ishaan Gulrajani, Tianyi Zhang, Yann Dubois, Xuechen Li, Carlos
  Guestrin, Percy Liang, and Tatsunori~B. Hashimoto.
\newblock Alpaca: A strong, replicable instruction-following model.
\newblock \emph{CRFM Stanford}, March 2023.
\newblock URL \url{https://crfm.stanford.edu/2023/03/13/alpaca.html}.

\bibitem[Touvron et~al.(2023)Touvron, Lavril, Izacard, Martinet, Lachaux,
  Lacroix, Rozière, Goyal, Hambro, Azhar, Rodriguez, Joulin, Grave, and
  Lample]{llama_paper}
Hugo Touvron, Thibaut Lavril, Gautier Izacard, Xavier Martinet, Marie-Anne
  Lachaux, Timothée Lacroix, Baptiste Rozière, Naman Goyal, Eric Hambro,
  Faisal Azhar, Aurelien Rodriguez, Armand Joulin, Edouard Grave, and Guillaume
  Lample.
\newblock Llama: Open and efficient foundation language models, 2023.

\bibitem[Wang et~al.(2023{\natexlab{a}})Wang, Ma, Feng, Zhang, Yang, Zhang,
  Chen, Tang, Chen, Lin, Zhao, Wei, and Wen]{survey_on_llm_agents}
Lei Wang, Chen Ma, Xueyang Feng, Zeyu Zhang, Hao Yang, Jingsen Zhang, Zhiyuan
  Chen, Jiakai Tang, Xu~Chen, Yankai Lin, Wayne~Xin Zhao, Zhewei Wei, and
  Ji-Rong Wen.
\newblock A survey on large language model based autonomous agents,
  2023{\natexlab{a}}.

\bibitem[Wang et~al.(2023{\natexlab{b}})Wang, Ivison, Dasigi, Hessel, Khot,
  Chandu, Wadden, MacMillan, Smith, Beltagy, and Hajishirzi]{camel_paper}
Yizhong Wang, Hamish Ivison, Pradeep Dasigi, Jack Hessel, Tushar Khot,
  Khyathi~Raghavi Chandu, David Wadden, Kelsey MacMillan, Noah~A. Smith,
  Iz~Beltagy, and Hannaneh Hajishirzi.
\newblock How far can camels go? exploring the state of instruction tuning on
  open resources, 2023{\natexlab{b}}.

\bibitem[Wei et~al.(2022)Wei, Zhang, Zhang, Gong, Zhang, Zhang, Yu, and
  Liu]{outlier_suppression_pushing_the_limit_paper}
Xiuying Wei, Yunchen Zhang, Xiangguo Zhang, Ruihao Gong, Shanghang Zhang,
  Qi~Zhang, Fengwei Yu, and Xianglong Liu.
\newblock Outlier suppression: Pushing the limit of low-bit transformer
  language models.
\newblock In Alice~H. Oh, Alekh Agarwal, Danielle Belgrave, and Kyunghyun Cho
  (eds.), \emph{Advances in Neural Information Processing Systems}, 2022.
\newblock URL \url{https://openreview.net/forum?id=yW5zeRSFdZ}.

\bibitem[Wolf et~al.(2020)Wolf, Debut, Sanh, Chaumond, Delangue, Moi, Cistac,
  Rault, Louf, Funtowicz, Davison, Shleifer, von Platen, Ma, Jernite, Plu, Xu,
  Le~Scao, Gugger, Drame, Lhoest, and Rush]{huggingface_paper}
Thomas Wolf, Lysandre Debut, Victor Sanh, Julien Chaumond, Clement Delangue,
  Anthony Moi, Pierric Cistac, Tim Rault, Remi Louf, Morgan Funtowicz, Joe
  Davison, Sam Shleifer, Patrick von Platen, Clara Ma, Yacine Jernite, Julien
  Plu, Canwen Xu, Teven Le~Scao, Sylvain Gugger, Mariama Drame, Quentin Lhoest,
  and Alexander Rush.
\newblock Transformers: State-of-the-art natural language processing.
\newblock In \emph{Proceedings of the 2020 Conference on Empirical Methods in
  Natural Language Processing: System Demonstrations}, pp.\  38--45, Online,
  October 2020. Association for Computational Linguistics.
\newblock \doi{10.18653/v1/2020.emnlp-demos.6}.
\newblock URL \url{https://aclanthology.org/2020.emnlp-demos.6}.

\bibitem[Wu et~al.(2023)Wu, Yao, and He]{ZeroQuant-FP_paper}
Xiaoxia Wu, Zhewei Yao, and Yuxiong He.
\newblock Zeroquant-fp: A leap forward in llms post-training w4a8 quantization
  using floating-point formats, 2023.

\bibitem[Xiao et~al.(2023)Xiao, Lin, Seznec, Wu, Demouth, and
  Han]{smoothquant_paper}
Guangxuan Xiao, Ji~Lin, Mickael Seznec, Hao Wu, Julien Demouth, and Song Han.
\newblock {S}mooth{Q}uant: Accurate and efficient post-training quantization
  for large language models.
\newblock In Andreas Krause, Emma Brunskill, Kyunghyun Cho, Barbara Engelhardt,
  Sivan Sabato, and Jonathan Scarlett (eds.), \emph{Proceedings of the 40th
  International Conference on Machine Learning}, volume 202 of
  \emph{Proceedings of Machine Learning Research}, pp.\  38087--38099. PMLR,
  23--29 Jul 2023.
\newblock URL \url{https://proceedings.mlr.press/v202/xiao23c.html}.

\bibitem[Yao et~al.(2023)Yao, Wu, Li, Youn, and He]{ZeroQuant-V2_paper}
Zhewei Yao, Xiaoxia Wu, Cheng Li, Stephen Youn, and Yuxiong He.
\newblock Zeroquant-v2: Exploring post-training quantization in llms from
  comprehensive study to low rank compensation, 2023.

\bibitem[Yu \& Wu(2023)Yu and
  Wu]{Compressing_transformers_features_are_lowrank_but_weights_are_not}
Hao Yu and Jianxin Wu.
\newblock Compressing transformers: Features are low-rank, but weights are not!
\newblock \emph{Proceedings of the AAAI Conference on Artificial Intelligence},
  37\penalty0 (9):\penalty0 11007--11015, Jun. 2023.
\newblock \doi{10.1609/aaai.v37i9.26304}.
\newblock URL \url{https://ojs.aaai.org/index.php/AAAI/article/view/26304}.

\bibitem[Yuan et~al.(2023)Yuan, Niu, Liu, Liu, Wang, Shang, Sun, Wu, Wu, and
  Wu]{rptq_paper}
Zhihang Yuan, Lin Niu, Jiawei Liu, Wenyu Liu, Xinggang Wang, Yuzhang Shang,
  Guangyu Sun, Qiang Wu, Jiaxiang Wu, and Bingzhe Wu.
\newblock Rptq: Reorder-based post-training quantization for large language
  models, 2023.

\bibitem[Zhang et~al.(2023)Zhang, Zhang, Shi, Chu, and
  Li]{LORA-FA:_MEMORY-EFFICIENT}
Longteng Zhang, Lin Zhang, Shaohuai Shi, Xiaowen Chu, and Bo~Li.
\newblock Lora-fa: Memory-efficient low-rank adaptation for large language
  models fine-tuning, 2023.

\bibitem[Zhu et~al.(2023)Zhu, Li, Liu, Ma, and Wang]{llm_compression_survey}
Xunyu Zhu, Jian Li, Yong Liu, Can Ma, and Weiping Wang.
\newblock A survey on model compression for large language models, 2023.

\end{thebibliography}
\bibliographystyle{iclr2021_conference}

% \appendix
% \section{Appendix: SVD Baseline}

% SVD Baseline results plots for largest CodeGen and StarCoder models.

% \section{Appendix: Inference Speedup for LoRDCoder.}

% \begin{table}
% \centering
% \begin{tabular}{|c|c|c|c|c|c|c|}
% \hline
% Model                                  & \multicolumn{2}{|c|}{1024 Tokens} & \multicolumn{2}{|c|}{512 Tokens } \\
%                                        & Time (ms) & Reduction & Time (ms) & Reduction \\
% \hline
% StarCoder 16B                          & 161.39    & N/A       & 78.49      & 1x \\
% \hspace{15mm}+\methodnameshort{} 14.9B & 154.89    & 4.03\%    & 73.55      & 0.0x \\
% \hspace{15mm}+\methodnameshort{} 14.5B & 149.17    & 7.57\%    & 70.89      & 0.0x \\
% \hspace{15mm}+\methodnameshort{} 13.8B & 149.02    & 7.66\%    & 72.95      & \\
% \hspace{15mm}+\methodnameshort{} 13.2B & 139.16    & 13.77\%   & 69.75      & \\
% \hspace{15mm}+\methodnameshort{} 12.6B & 131.55    & 18.49\%   & 60.95      & \\
% \hspace{15mm}+\methodnameshort{} 12.3B & 128.64    & 20.29\%   & 61.51      & \\
% \hline
% \end{tabular}
% \caption{\label{tab:metrics} Performance Metrics}
% \end{table}

% \section{Appendix: Extending to General Purpose LLMs}

\end{document}